\documentclass[]{article}
\usepackage{arxiv}
\pdfoutput=1

\usepackage[T1]{fontenc}
\usepackage[english]{babel}
\usepackage[numbers,compress]{natbib}
\usepackage[utf8]{inputenc}
\usepackage{amsmath}
\usepackage{hyperref}

\usepackage[dvipsnames]{xcolor}
\usepackage{amsfonts}
\usepackage{amssymb}
\usepackage{amsthm}
\usepackage{booktabs}
\usepackage{braket}
\usepackage{comment}
\usepackage{enumitem}
\usepackage{multicol}
\usepackage{multirow}
\usepackage{pifont}
\usepackage{url}
\usepackage{graphicx}
\usepackage{authblk}

\newcommand{\cmark}{\textcolor{ForestGreen}{\ding{51}}}
\newcommand{\xmark}{\textcolor{red}{\ding{55}}}
\newcommand{\init}{{\textsf{init}}}

\newcommand{\halt}{{\textsf{halt}}}
\newcommand{\non}{{\textsf{non}}}
\newcommand{\acc}{{\textsf{acc}}}
\newcommand{\rej}{{\textsf{rej}}}
\newcommand{\sos}{{\#}}
\newcommand{\eos}{{\$}}
\newcommand{\mat}[1]{{\mathbf{#1}}}
\newcommand{\QFL}{{\mathsf{QFL}}}
\newcommand{\MMQFL}{{\mathsf{MMQFL}}}
\newcommand{\MOQFL}{{\mathsf{MOQFL}}}
\newcommand{\coMMQFL}{{\mathsf{co\text{-}MMQFL}}}
\newcommand{\coMOQFL}{{\mathsf{co\text{-}MOQFL}}}
\newcommand{\co}{{\mathsf{co}}}
\newcommand{\subend}{{\mathsf{end}}}
\newcommand{\subcoend}{{\mathsf{coend}}}
\newcommand{\MOD}{{\mathsf{MOD}}}
\newcommand{\EQU}{{\mathsf{EQU}}}
\newcommand{\MAPE}{{\text{MAPE}}}

\newcommand{\R}{{\mathbb{R}}}
\newcommand{\C}{{\mathbb{C}}}

\DeclareMathOperator*{\argmax}{{arg\,max}}
\DeclareMathOperator*{\argmin}{{arg\,min}}

\newtheorem{theorem}{Theorem}
\newtheorem{property}{Property}
\newtheorem{definition}{Definition}
\newtheorem{proposition}{Proposition}
\newtheorem{corollary}{Corollary}

\begin{document}

\title{A Framework for Quantum Finite-State Languages with Density Mapping}


\renewcommand*{\Authfont}{\bfseries\large}
\renewcommand*{\Affilfont}{\rm\normalsize}
\renewcommand\Authsep{\quad\quad}
\renewcommand\Authand{\quad\quad}
\renewcommand\Authands{\quad\quad}

\author{SeungYeop Baik}
\author{Sicheol Sung}
\author{Yo-Sub Han}
\affil{
    Yonsei University \protect\\
    Seoul, Republic of Korea \protect\\
    \texttt{\{sybaik2006,sicheol.sung,emmous\}@yonsei.ac.kr}
}
\maketitle
\rhead{}

\begin{abstract}
A quantum finite-state automaton (QFA) is a theoretical model designed to
simulate the evolution of a quantum system with finite memory in response to
sequential input strings. We define the language of a QFA as the set of strings
that lead the QFA to an accepting state when processed from its initial state.
QFAs exemplify how quantum computing can achieve greater efficiency compared to
classical computing. While being one of the simplest quantum models, QFAs are
still notably challenging to construct from scratch due to the preliminary
knowledge of quantum mechanics required for superimposing unitary constraints on
the automata. Furthermore, even when QFAs are correctly assembled, the
limitations of a current quantum computer may cause fluctuations in the
simulation results depending on how an assembled QFA is translated into a
quantum circuit. 

We present a framework that provides a simple and intuitive way to build QFAs
and maximize the simulation accuracy. Our framework relies on two  methods:
First, it offers a predefined construction for foundational types of QFAs that
recognize  special languages~$\MOD_n$ and $\EQU_k$. They play a role of basic
building blocks for more complex QFAs. In other words, one can obtain more
complex QFAs from these foundational automata using standard language
operations. Second, we improve the simulation accuracy by converting these QFAs
into quantum circuits such that the resulting circuits perform well on noisy
quantum computers.
\end{abstract}


\section{Introduction}

The scientific field of quantum mechanics inspired an advent of quantum
computational models in the 1980s~\cite{Benioff80}. These models have enabled
the development of quantum algorithms designed to tackle problems that are
infeasible on classical computers, thus lending to their advantage. Investment
in quantum computing has exploded with the recent development of physical
quantum computers. With greater funding, research on quantum resource
optimization has been explored across multiple fields, such as chemistry,
quantum simulation and computational
theory~\cite{EmuCS23,ArrazolaDBL20,HaganW23,King23}.

The intrinsic requirements of quantum algorithms propagate to the quantum
computers on which they run. These requirements include the number of qubits,
the accuracy of gates and topological structures. Currently, quantum computers
fail to meet these requirements and rely on error suppression and error
mitigation, both of which have implicit cost. Previous works have examined
computational models with
restrictions.~\cite{NishimuraO09,Nishimura03,GainutdinovaAYA18,VieiraB22,HuangRGN23}.
We consider to use quantum finite-state automata (QFAs) that represent quantum
computers with the restriction of limited qubits.

The concept of QFAs has been present since the inception of quantum
computation~\cite{KondacsW97}, and extensive theoretical research has been
conducted in this area~\cite{BrodskyP02, BertoniMP03, AmbainisANATAVU99,
GiannakisKPCAT15}. However, the realization of QFAs as quantum circuits has only
recently emerged with the introduction of functional quantum
computers~\cite{BirkanSONY21}. This research area is still in its early stages,
with researchers facing two main challenges: the difficulty of composing QFAs
and the inaccuracies of quantum computers.

Constructing QFAs from scratch is challenging for those without knowledge of
quantum mechanics or formal language theory. This complexity arises from the
constraints on transitions in QFAs, where transitions must adhere to the
properties of a unitary matrix. Properly constructed QFAs are still prone to
obtaining practical results that deviate significantly from theoretical ones,
due to the high error rate associated with quantum computers. This discrepancy
is attributed to errors introduced during computation, which can alter the
bit-representation of each state in the QFA circuit.

\begin{figure*}
\centering
\includegraphics[width=\textwidth]{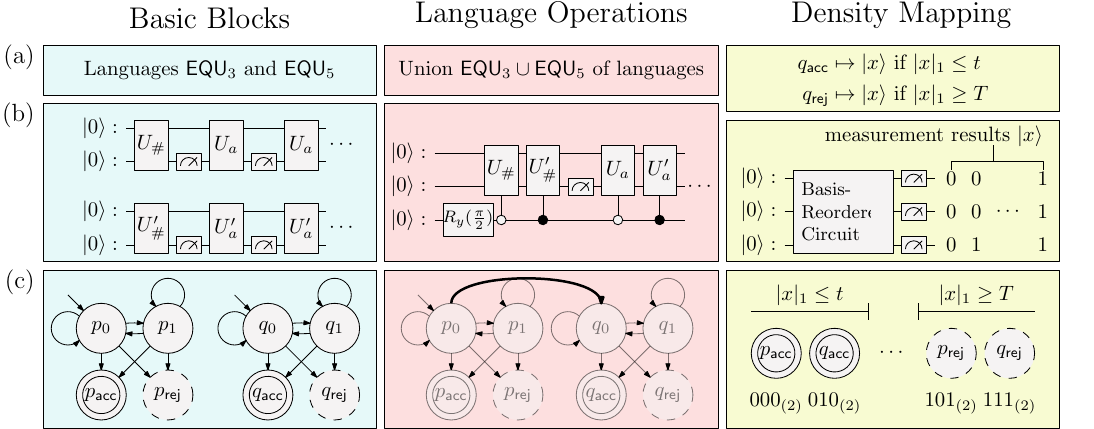}
\caption{
The three main features of our framework are as follows: basic blocks, language
operations and density mapping. Each feature is presented in (a)~a formal
description, (b)~a circuit and (c)~a graphical representation.
}
\label{fig:scenario}
\end{figure*}

We propose a framework that addresses these challenges and enhances simulation
accuracy by providing intuitive QFA composition methods alongside improvements
to the transpilation process from QFAs to quantum circuits. Our framework
enables users to compose complex QFAs from simple QFAs using language
operations. Our proposed methodology extends the notable frameworks of
classical automata, such as FAdo~\cite{AlmeidaAAMR09}, FSA~\cite{KanthakN04} and
others~\cite{Mousavi16,MohriPR97,AllauzenRSSM07,MayK06,ArrivaultBDE17,Brainerd69}.
In our proposed framework, we also introduce the novel concept of \emph{density
mapping}, a method employed in our framework for determining the
bit-representation of each state in a QFA during transpilation. Density mapping
is designed to reduce the probability of fluctuation caused by bit-flip errors.
Fig.~\ref{fig:scenario} demonstrates how a user can compose and simulate quantum
circuits using the proposed framework. The intended workflow of our proposed
framework is threefold: (1)~constructing simple QFAs, (2)~combining these simple
QFAs with language operations to compose a \emph{complex} QFA and
(3)~transpiling the resulting complex QFA into a circuit using density mapping.
The given workflow lends to a more intuitive approach to quantum circuit
construction as well as improved simulation accuracy therein.

In summary, we propose a framework that facilitates the construction and
accurate simulation of QFAs. Our main contributions are as follows.
\begin{enumerate}
\item
We introduce a novel QFA construction of unary finite languages,
which serves as the elementary units of our framework.
\item
We prove closure properties of various language classes defined by QFAs to
illustrate how our framework combines QFAs.
\item
We propose the $\mathcal{D}$-mapping technique to enhance the QFA simulation
accuracy on quantum computers.
\end{enumerate}

The rest of the paper is organized as follows. Section~\ref{sec:prelim} provides
a brief background on QFAs and their languages. In Section~\ref{sec:framework},
we present details of our framework. We demonstrate the effectiveness of our
density mapping in Section~\ref{sec:experiments} by running experiments on the
proposed framework. Finally, we summarize the results and outline possible
future work in Section~\ref{sec:conclusion}. Our framework is publicly
accessible at~\url{https://github.com/sybaik1/qfa-toolkit}.

\section{Preliminaries}

\label{sec:prelim}
We first describe notations and definitions of two types of QFA: one is a
measure-once (MO-) QFA and the other is a measure-many (MM-) QFA. We then
define several types of languages recognized by MO-QFAs and MM-QFAs.

\subsection{Basic Notations}

Symbols~$\mathbb{R}$ and $\mathbb{C}$ denote the set of real numbers and the set
of complex numbers, respectively. For a finite set~$Q$ with $n$ elements,
$\C^{Q}$ and $\C^{Q \times Q}$ denote the set of column vectors of dimensions
$n$ and the set of matrices of sizes $(n, n)$ whose components are indexed by
elements of $Q$ and $Q \times Q$, respectively. For an element $a \in Q$,
$\ket{a} \in \C^Q$ is a unit vector whose components are all zero, except an
$a$-indexed component. The set~$\{\ket{a} \mid a \in Q\}$ is a \emph{standard
basis} of the vector space~$\C^{Q \times 1}$. A vector~$\psi \in \C^{Q \times
1}$ has a unique expression as a linear combination:
\[
    \psi = \sum_{a \in Q} \alpha_a \cdot \ket{a},
    \text{where $\alpha_a \in \C$ for all $a \in Q$.}
\]
Then, $|\psi| := \sqrt{\sum_{a \in Q}|\alpha_a|^2}$ denotes the \emph{length}
of~$\psi$.

We use $\mat M^T, \mat M^{-1}$, $\overline{\mat M}$ and $\mat M^\dag$ to denote
a \emph{transpose}, \emph{inverse}, \emph{conjugate} and
\emph{conjugate-transpose} of a matrix $\mat M$, respectively. $\det \mat M$
denotes the \emph{determinant} of a given matrix~$\mat M$. A matrix~$\mat U$ is
considered \emph{unitary} if it satisfies the equality: $\mat U^{-1} = \mat
U^\dag$. Given a finite set $A$ of indices and a set $S \subseteq A$, a
projection matrix~$\mat P_S$ onto $S$ is a zero-one diagonal matrix where $(\mat
P_S)_{(a, a)} = 1$ if $a \in S$ and $(\mat P_S)_{(a, a)} = 0$, otherwise. For
matrices~$\mat M_1 \in \C^{A \times A}$ and $\mat M_2 \in \C^{B \times B}$ with
$A \cap B = \emptyset$, their \emph{direct sum} is
\[
\mat M_1 \oplus \mat M_2 = \begin{bmatrix}
    \mat M_1 & \mat 0 \\
    \mat 0 & \mat M_2
\end{bmatrix} \in \C^{(A \cup B) \times (A \cup B)}.
\]

A \emph{superposition} of $n$ qubits is formalized as a $2^n$-dimensional
complex vector with a length of 1, and its evolution is described by a unitary
matrix. The basis vectors of the superposition are represented by
$\ket{00\cdots0}, \ket{10\cdots0}, \ldots, \ket{11\cdots1}$, where each digit
corresponds to the value of each qubit. The measurement of the superposition is
formalized by an \emph{observable}, which is a partitioning~$A_1 \cup \cdots
\cup A_k$ of the dimensions. When a superposition~$\ket{\psi}$ is measured with
an observable, it collapses to one of the vectors~$\ket{\psi'_i} :=
P_{A_i}\ket{\psi}$, each with a probability $|\ket{\psi'_i}|^2$.

\subsection{Quantum Finite-State Automata and Languages}

A finite-state automaton (FA) is a model that simulates the behavior of a system
changing its state between a fixed number of distinct states. The automaton
updates its state in an online manner in response to external
inputs---formalized as strings over a finite alphabet. A quantum finite-state
automaton (QFA) is a quantum variant of FA, which can exist in a superposition
comprised of a finite number of states. We consider two different types of QFAs:
an MO-QFA, which is observed only at the end of sequential
inputs~\cite{MooreC00}, and an MM-QFA, which is observed after every step of the
sequence~\cite{KondacsW97}.

\begin{figure*}
\centering
\includegraphics[width=0.8\textwidth]{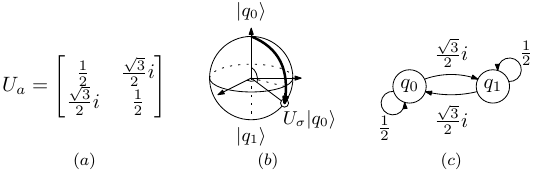}
\caption{
The relation between
(a)~a unitary transition matrix of a semi-QFA,
(b)~a change of superposition on the Bloch sphere according to a transition and
(c)~a graphical representation of the matrix's transitions.
Transitions of QFAs, for example $\ket{q_0} \mapsto U_a\ket{q_0}$, are
represented as a unitary transformation in Hilbert space.
}
\label{fig:qfa_repr}
\end{figure*}

Let $\Sigma$ denote a finite \emph{input alphabet}, which consists of the
symbols used in sequential input. A \emph{semi-QFA} then describes the evolution
of a QFA for each symbol. Fig.~\ref{fig:qfa_repr} shows how transitions of a
semi-QFA are represented graphically.

\begin{definition}
A semi-QFA is a tuple~$(Q, \Sigma,
\{\mat{U}_\gamma\}_{\gamma \in \Gamma})$, where
(1)~$Q$ is a finite set of \emph{state},
(2)~$\{\mat{U}_\gamma\}_{\gamma \in \Gamma}$ is the unitary matrix for each
symbol $\gamma\in \Gamma$ and
(3)~$\Gamma := \Sigma \cup \{\sos, \eos\}$ is the \emph{tape alphabet}, with
$\sos$ and $\eos$ as the start-of-string and end-of-string symbols,
respectively.
\end{definition}

We can define MO-QFAs and MM-QFAs as semi-QFAs with two defining distinctions:
the inclusion of an initial state as well as that of final states, which are
either accepting or rejecting.

\begin{definition}
An \emph{MO-QFA} is specified by a tuple\footnote{In some other papers, the tape
alphabet of an MO-QFA does not include the end-of-string symbol. This does not
affect the expressive power or the size of MO-QFAs~\cite{BrodskyP02}.}~$(Q,
\Sigma, \{\mat{U}_\gamma\}_{\gamma \in \Gamma}, q_\init, Q_\acc)$, where
(1)~a triple~$(Q, \Sigma, \{\mat{U}_\gamma\}_{\gamma \in \Gamma})$ is a semi-QFA;
(2)~$q_\init \in Q$ is an \emph{initial state}; and
(3)~$Q_\acc \subseteq Q$ is a set of \emph{accepting states}.
\end{definition}

An MO-QFA reads each symbol of string one by one and updates its superposition
of states at each based on their transition matrix corresponding to the symbol.
After reading the entire string, the MO-QFA measure the the superposition of
states. It accepts the string if and only if the measured state is an accepting
state; otherwise, it reject the string.

Formally, for a string~$x = x_1 x_2 \cdots x_n \in \Gamma^*$ on the tape
alphabet, we denote the unitary matrix~$\mat U_{x_n} \mat U_{x_{n-1}} \cdots
\mat U_{x_1}$ as $\mat U_{x}$. We also denote $\mat P_{Q_\acc}$, $\mat
P_{Q_\rej}$ and $\mat P_{Q_\non}$ as $\mat P_\acc$, $\mat P_\rej$ and $\mat
P_\non$, respectively, for simplicity of notation. Then, the probability of an
MO-QFA~$M$ accepting string~$w$ is $|\mat P_\acc \ket{\psi}|^2$, where
$\ket{\psi}$ is the superposition~$\mat U_{\sos w \eos} \ket{q_\init}$ of the
QFA after reading the end-of-string symbol.

\begin{definition}
An \emph{MM-QFA} is specified by a tuple~$(Q, \Sigma, \{\mat{U}_\gamma\}_{\gamma
\in \Gamma}, q_\init, Q_\acc, Q_\rej)$, where (1)~$(Q, \Sigma,
\{\mat{U}_\gamma\}_{\gamma \in \Gamma})$ is a semi-QFA; (2)~$q_\init \in Q$ is
an \emph{initial state}; (3)~$Q_\acc, Q_\rej \subseteq Q$ are sets of
\emph{accepting} and \emph{rejecting states}, respectively. We denote a set of
\emph{non-halting} states~$Q_\non :=  Q \setminus (Q_\acc \cup Q_\rej)$
\end{definition}

The MM-QFA reads the tape symbols one by one for a given string and updates the
superposition according to the transition matrix, similar to MO-QFAs. However,
unlike MO-QFAs, it measures their superposition at the end of each step, causing
it to collapse onto one of three sub-spaces $\C^{Q_\acc}$, $\C^{Q_\rej}$ or
$\C^{Q_\non}$. The MM-QFA then accept or reject the string immediately if the
resulting superposition is on $\C^{Q_\acc}$ the MM-QFA~$M$or $\C^{Q_\rej}$,
respectively. Otherwise, it keeps reading the next symbol.

Kondacs and Watrous~\cite{KondacsW97} introduced \emph{total states} to
formalize the decision process of MM-QFAs. A total state is a triple~$(\psi,
p'_\acc, p'_\rej) \in \C^Q \times \R \times \R$, where $p'_\acc$ and $p'_\rej$
are the accumulated probabilities of acceptance and rejection. The vector~$\psi$
denotes the superposition and the probability of non-halting by its direction
and the square of its length, respectively.
For an MM-QFA~$M$, $T_\gamma$ is an evolution function of the internal state of
$M$ with respect to $\gamma$ defined as follows:
\begin{align*}
    T_\gamma&: (\psi, p'_\acc, p'_\rej)
    \mapsto (
        \mat P_\non \mat U_\gamma \psi,
        p'_\acc + \Delta p'_\acc,
        p'_\rej + \Delta p'_\rej
    ),
\end{align*}
where
\[
    \Delta p'_\acc := | \mat P_\acc \mat U_\gamma \psi |^2
    \text{ and }
    \Delta p'_\rej := | \mat P_\rej \mat U_\gamma \psi |^2.
\]
For a string~$x = x_1 x_2 \cdots x_n \in \Gamma^*$ on the tape alphabet, we use
$T_{x}$ to denote $T_{x_n} \circ T_{x_{n-1}} \circ \cdots \circ T_{x_1}$. Then,
$(\psi, p_\acc, p_\rej) := T_{\sos w \eos} (\ket {q_\init}, 0, 0)$ denotes the
final total state, and $p_\acc$ is the probability that $M$ accepts a given
string~$w$.

We say that an MM-QFA~$M$ is \emph{valid} if $p_\acc + p_\rej = 1$ for any $w
\in \Sigma^*$. If $p_\acc = 0$ for every~$w \in \Sigma^*$ such that $(\_,
p_\acc, \_) := T_{\sos w} (\ket {q_\init}, 0, 0)$, then we say that $M$ is
\emph{end-decisive}. In other words, $M$ does not accept strings before reading
the end-of-string symbol. Similarly, $M$ is \emph{co-end-decisive} if it only
rejects strings when reading the end-of-string symbol~\cite{BrodskyP02}. For the
rest of the paper, we focus solely on valid MM-QFAs.

Let $M$ be an MO-QFA or an MM-QFA and $f_M(w)$ denote the probability that a given $M$
accepts $w$. We call the function~$f_M: \Sigma^* \to \mathbb{R}$ the
\emph{stochastic language} of $M$. We say an MO- or MM-QFA~$M$ recognizes $L$
with a \emph{positive one-sided} error of margin~$\varepsilon > 0$ if
(1)~$f_M(w) = 0$ for $w \notin L$ and (2)~$f_M(w) > \varepsilon$. In other
words, $M$ always rejects $w \notin L$ and accepts $w \in L$ with
probability~$f_M(w) > \varepsilon$. Similarly, we say $M$ recognizes $L$ with a
\emph{negative} one-sided error of margin~$\varepsilon > 0$ if (1)~$f_M(w) = 1$
for $w \in L$ and (2)~$f_M(w) < 1 - \epsilon$ for $w \notin L$.

We define the following classes of \emph{quantum finite-state
languages~(QFLs)}~$\MOQFL$, $\MMQFL$, $\MMQFL_\subend$ and $\MMQFL_\subcoend$ as
those recognized by MO-QFAs, MM-QFAs, end-decisive MM-QFAs and co-end-decisive
MM-QFAs, respectively. We also define the class of complements for each language
family, which are denoted using a prefix~$\co$-. For example, $\coMOQFL$ denotes
the class~$\left\{ L \subseteq \Sigma^* \mid \overline{L} \in \MOQFL \right\}$.

\subsection{Closure Properties of Stochastic Languages of QFAs}

Stochastic languages of QFAs possess known closure properties, specific to
MO-QFAs under the following
operations~\cite{MooreC00,BertoniMP03,AmbainisAKAVM01}.
\begin{enumerate}[align=left]
\item[(Hadamard product)]
$(f_M \otimes f_N)(w) = f_M(w) \cdot f_N(w)$
\item[(Convex linear combination)]
For $c_1, c_2 \in [0, 1]$ with $c_1 + c_2 = 1$,
\[ (c_1 f_M \oplus c_2 f_N)(w) = c_1 \cdot f_M(w) + c_2 \cdot f_N(w) \]
\item[(Complement)] $\overline{f_M}(w) = 1 - f_M(w)$
\item[(Inverse homomorphism)]
$(h^{-1} \circ f_M)(w) = f_M(h^{-1}(w))$
\end{enumerate}

Compared to the stochastic languages of MO-QFAs, those of MM-QFAs are closed
under linear combination, complement and inverse
homomorphism~\cite{BertoniMP03,BianchiP10} but not under the Hadamard product.
For the cases of end-decisive and co-end-decisive MM-QFAs, their stochastic
languages are closed under linear combination and inverse homomorphism. However,
end-decisive MM-QFAs, in particular, are also closed under the Hadamard
product~\cite{BianchiP10}. Table~\ref{tab:qfa_closure_property} summarizes the
closure properties of stochastic languages of QFAs. We also present a
construction for the convex linear combination to complete the paper.

\begin{table}
\caption{
The closure properties of stochastic languages~$f$ and $g$ for each class of
QFAs. Check mark~(\cmark) and Cross mark~(\xmark) denotes whether the class is
closed under the operation or not.
}
\label{tab:qfa_closure_property}
\centering
\begin{tabular}{lcccc}
\toprule
QFA Class
& $f_M \otimes f_N$
& $c_1 f_M \oplus c_2 f_N$
& $\overline{f_M}$
& $h^{-1} \circ f_M$ \\
\midrule
MO-QFA
& \cmark & \cmark & \cmark & \cmark \\
MM-QFA
& \xmark & \cmark & \cmark & \cmark \\
End-dec.
& \cmark & \cmark &  & \cmark\\
Co-end-dec.
& & \cmark &   & \cmark \\
\bottomrule
\end{tabular}
\addtolength{\tabcolsep}{-3pt}  
\end{table}

\begin{proposition}
Consider two MM-QFAs
\begin{enumerate}
    \item 
$M := \left(
    Q, \Sigma, \{\mat U_\gamma \}_{\gamma \in \Gamma}, q_\init, Q_\acc, Q_\rej
\right)$ and
\item 
$N := \left(
    P, \Sigma, \{\mat V_\gamma \}_{\gamma \in \Gamma}, p_\init, P_\acc, P_\rej
\right).$
\end{enumerate}
For $c_1, c_2 \in [0, 1]$ with $c_1 + c_2 = 1$,
the following MM-QFA~$c_1 M \oplus c_2 N$
satisfies $(c_1 M \oplus c_2 N)(w) = c_1 M(w) + c_2 N(w)$.
\[
c_1 M \oplus c_2 N := ( Q \cup P, \Sigma, \{\mat W_\gamma\}_{\gamma \in \Gamma},
q_\init, Q_\acc \cup P_\acc, Q_\rej \cup P_\rej),
\]
where
$\mat W_\gamma := \mat U_\gamma \oplus \mat V_\gamma$ for $\gamma \in \Sigma
\cup \{ \eos \}$ and
\[
\mat W_\sos := \left(
\mat U_\sos \oplus \mat V_\sos \right) \cdot
    \begin{bmatrix}\begin{array}{cccc}
       \sqrt{c_1} & \mat 0 & i \sqrt{c_2}& \mat 0 \\
       \mat 0 & \mat I & \mat 0 & \mat 0 \\
       i \sqrt{c_2} & \mat 0 & \sqrt{c_1} & \mat 0 \\
       \mat 0 & \mat 0 & \mat 0 & \mat I \\
    \end{array}
    \end{bmatrix}
    \in \C^{(Q \cup P) \times (Q \cup P)}.
\]
If $M_1$ and $M_2$ are both end-decisive or co-end-decisive,
then the resulting~$M$ is also end-decisive or co-end-decisive, respectively.
\end{proposition}

\section{Proposed Framework for QFLs}
\label{sec:framework}

It is more complex to construct QFAs as opposed to classic FAs due to the
unitary constraints that a QFA must preserve. On top of the difficulties with
QFAs, when constructing a quantum circuit that recognizes a QFL, one has to
consider a margin with the accepting probability and rejecting probability---the
accepting policy~\cite{BrodskyP02,AmbainisF98}. Several accepting policies
exist, such as one-sided errors, two-sided errors, isolated cut-points,
threshold acceptance and threshold with margin~\cite{BertoniMP03}. In our
framework, we only consider one-sided errors since they offer a compromise
between the language coverage and the closure properties.

When designing a quantum circuit for a language, a typical process of making
their QFA proceeds as follows: First, determine the number of states for the
target QFA. Then, construct transition matrices for each of the states adhering
to unitary constraints. Subsequently, verify if the created QFA can
differentiate between accepting and rejecting strings. If it fails to do so,
adjustments to the transitions design or an increase in the number of states are
necessary. If the target QFA functions as intended, it can be transpiled into a
quantum circuit.

The aforementioned process is not the sole paradigm of designing a circuit, but
the continuous evolution of which, boasting both advancements and errors, are
necessary in the development of the target quantum circuit. We overcome these
errors and inconveniences by encapsulating the design process through the use of
QFLs as building blocks. By doing so, we can employ operations between the
building blocks to construct the desired QFL. In our framework, we provide
constructions for the two languages $\MOD_p$ and $\EQU_k$ as building blocks,
leveraging them to create more complex and diverse languages. In the above
scenario, instead of deciding on the size and transitions, we can decompose the
language into a set of expressions with $\MOD_p$ and $\EQU_k$, then compose them
to get the QFA. We can finally transpile the QFA with our density mapping to get
the target quantum circuit.

\subsection{Basic Blocks---Two QFLs with Simple Constructions}

We settled on small and simple QFAs as the building blocks of our framework and
give the characteristics and properties of two QFAs $\MOD_n$ and $\EQU_k$.

First, we revisit $\MOD_n:= \{a^j \mid j \equiv 0 \mod n \}$ that has several
previous approaches due to its small and simple circuit~\cite{BirkanSONY21}. It
is known that there exists a two-state MO-QFA that recognizes the
language~$\MOD_n$ with a negative one-sided bounded error. Furthermore, for each
prime~$p$, there exists an $O(\log p)$-state MO-QFA recognizing $\MOD_p$ with an
error margin of $1/8$, which is independent of the choice of $p$. The circuit of
$\MOD_n$ and $\MOD_p$ requires only one and $O(\log {\log p})$ qubits,
respectively~\cite{AmbainisF98}. Next, we present our construction of an MM-QFA
for $\EQU_k := \{a^k\}$.

\begin{theorem}\label{theorem:equ_construction}
Let $\{a\}$ be a unary alphabet. For a singleton language~$\EQU_k$, and
parameters $\theta, \varphi$ with $0 < \theta, \varphi < \frac{\pi}{2}$, let
$M_k(\theta, \varphi)$ be an co-end-decisive MM-QFA defined as follows.
\[
    M_k := (Q, \{a\}, \{\mat{U}_\gamma\}_{\gamma \in \{\sos, a, \eos \}}, q_0,
        \{q_\acc\}, \{q_\rej\}),
\]
where $Q := \{q_0, q_1, q_\acc, q_\rej\}.$
The specific unitary operators are given by
{\small
\[
    \mat{U}_\sos := \begin{bmatrix}
        \cos \theta & -\sin \theta & 0 & 0 \\
        \sin \theta & \cos \theta & 0 & 0 \\
        0 & 0 & 1 & 0 \\
        0 & 0 & 0 & 1
    \end{bmatrix},
    \mat{U}_a := \begin{bmatrix}
        \cos \varphi & 0 & -\sin \varphi & 0 \\
        0 & 1 & 0 & 0 \\
        \sin \varphi & 0 & \cos \varphi & 0 \\
        0 & 0 & 0 & 1
    \end{bmatrix}
    \text{ and }
    \mat{U}_\eos :=
    \sqrt{C_k}
    \cdot \begin{bmatrix}
        \mat{0} & \mat{A}_k^\dag \\
        \mat{A}_k & \mat{0}
    \end{bmatrix},
\]
}
where $\mat{A}_k$ is a sub-matrix given by
\[
    \mat{A}_k :=
    \begin{bmatrix}
        \cos^k \varphi \cdot \cos \theta & \sin \theta \\
        -\sin \theta & \cos^k \varphi \cdot \cos \theta
    \end{bmatrix},
\]
and the factor~$C_k$ is given by
$C_k := \cos^{2k} \varphi \cos^2 \theta + \sin^2 \theta.$
Then, $M_k(\theta, \varphi)$ recognizes the language~$\{a^k\}$
with a negative error bound with a margin~$\varepsilon$ satisfying
\[
    0 < \varepsilon < C_k^{-1} \cdot (\cos \theta \sin \theta \cos^k \varphi
    \cdot (1 - \cos \varphi))^2.
\]
\end{theorem}

\begin{proof}
Fix $M := M_k(\theta, \varphi)$ for some parameters~$k$, $\theta$ and $\varphi$.
The validity of $M$ is obvious from the construction. For each $j \ge 0$, let
$T_{\sos a^j} (\ket{q_0}, 0, 0) =: (\psi_j,p_{\acc, j},p_{\rej, j}).$ Then, the
values of components are $\psi_0 = \mat{P}_\non U_\sos \ket{q_0} = \cos \theta
\ket{q_0} + \sin \theta \ket{q_1}$ and $\psi_{j+1} = \mat{P}_\non U_a \psi_j$.
By induction, $\psi_j = \cos^k \varphi \cos \theta \ket{q_0} + \sin \theta
\ket{q_1}.$ Now we consider the final total state~$T_{\sos a^j \eos} (\ket{q_0},
0, 0) =: (\psi, p_\acc(j), p_\rej(j)).$ The rejecting probability of a
string~$a^j$ is
\begin{align*}
    p_\rej(j)
    &= |\mat{P}_\rej \mat{U}_\eos \psi_j|^2 = |\mat{P}_\rej
        \mat{U}_\eos (\cos^j \varphi \cos \theta \cdot \ket{q_0} + \sin \theta
        \cdot \ket{q_1})|^2 \\
    &= C_k^{-1} \cdot |(\cos^k \varphi -\cos^j \varphi) \cos \theta \sin \theta
    \cdot \ket{q_\rej}|^2,
\end{align*}
and $_M(a^k) = p_\acc(k) = 1 - p_\rej(k) = 1.$ From the observation that
$\argmin_{j \ne k} p_\rej(j) = k+1,$ we finally obtain
\begin{align*}
    f_M(a^j) &= 1 - p_\rej(j) \le 1 - p_\rej(k+1) = 1 - C_k^{-1} \cdot (\cos
        \theta \sin \theta \cos^k \varphi \cdot (1 - \cos \varphi))^2.
\end{align*}
\end{proof}

One might need to execute the QFA matching process multiple times to obtain a
correct result, as the probability of obtaining the correct results decreases
with the increasing length of the input string. The subsequent theorem describes
how many repetitions are required to achieve a predetermined probability of
accuracy.

\begin{theorem}\label{theorem:fixed_margin}
Fix the parameters~$\theta_0, \varphi_0$ and define $\alpha := \cos \theta_0$
and $\beta := \cos \varphi_0$. Let $M_k := M_k(\theta_0, \varphi_0)$ for each
$k$. Then, $(M_k(a^j))^N < \frac{1}{e} \approx 0.37$ for each $j \ne k$, where
\[
    N := \frac{1}{(1-\alpha^2)(1-\beta)^2}
        + \frac{1}{\alpha^2\beta^{2k}(1-\beta)^2} - 1 \in 2^{\Omega(k)}.
\]
\end{theorem}

\begin{proof}
Note that $a^j$ has the minimum reject probability when $j = k+1$,
except in the case where $j = k$.
We can compute the value of $M_k(a^{k+1})$ as follows:
\begin{align*}
   M_k(a^{k+1}) &= 1 - \frac{
        (\cos \theta \cdot \sin \theta \cdot \cos^k \varphi (1 - \cos
        \varphi))^2
    }{\cos^{2k} \varphi \cdot \cos^2 \theta + \sin^2 \theta} \\
    &= \frac{
        \alpha \beta^{2k} + (1-\alpha^2) - \alpha^2 (1 - \alpha^2)
        \beta^{2k}(1-\beta)^2
    }{\alpha^2 \beta^{2k} + (1-\alpha^2)} \\
    &= \left(1 + \frac{
        \alpha^2 (1 - \alpha^2) \beta^{2k}(1-\beta)^2
    }{
        \alpha \beta^{2k} + (1-\alpha^2) - \alpha^2 (1 - \alpha^2)
        \beta^{2k}(1-\beta)^2
    }\right)^{-1} \\
    &= \left(1
        + \left(\frac{1}{(1-\alpha^2)(1-\beta)^2}
        + \frac{1}{\alpha^2\beta^{2k}(1-\beta)^2} - 1\right)^{-1}\right)^{-1} \\
    &= \left(1 + \frac{1}{N}\right)^{-1}.
\end{align*}
Then, $M_k(a^j)^N \le M_k(a^{k+1})^N \le (1 + \frac{1}{N})^{-N} < \frac{1}{e}$
for $j \ne k$. Therefore, when $\theta$ and $\varphi$ are fixed independently of
$L_k$, one have to run $M_k(\theta, \varphi)$ on $w$ for $2^{O(k)}$-times to
decide $w \in L_k$ within a fixed probability.
\end{proof}

Note that Theorems~\ref{theorem:equ_construction} and~\ref{theorem:fixed_margin}
examine the condition where parameters~$\theta$ and $\varphi$ are chosen
independently to the target language~$L_k$. The following proposition describes
how our framework finds optimal parameters that maximize the error margin for
each $k$.

\begin{proposition}\label{prop:optimal_param}
Let $\omega$ be a unique positive solution of $\omega^{k+1} + (k+1) \omega - k =
0$. Then, parameters~$\theta = \tan^{-1}\sqrt{\omega^k}$ and $\varphi =
\cos^{-1}\omega$ maximize the error margin of $M_k(\theta, \varphi)$.
\end{proposition}

\begin{proof}
We want to maximize the rejection probability~$p_{\rej}(\theta, \varphi)$ of
$a^{k+1}$ where
\[
    p_\rej(\theta, \varphi) =  \frac{
        (\cos \theta \sin \theta \cos^k \varphi (1 - \cos \varphi))^2
    }{\cos^{2k} \varphi \cos^2 \theta + \sin^2 \theta}.
\]
Let $f(\theta, \varphi) := (p_\rej(\theta, \varphi))^{-1}$ be an objective
function. Then, the following holds.
\begin{align*}
    (\theta_k, \varphi_k)
        &:= \argmax_{\theta,\varphi} p_\rej(\theta, \varphi)
            = \argmin_{\theta,\varphi} f(\theta, \varphi) \\
        &= \argmin_{\theta,\varphi} \frac{1}{\sin^2 \theta (1 - \cos \varphi)^2}
            + \frac{1}{\cos^2 \theta \cos^{2k} \varphi (1 - \cos \varphi)^2}
\end{align*}
Since $f$ is differentiable on the domain $0 < \theta, \varphi < \frac{\pi}{2}$,
the equation $(\partial_\theta f)(\theta_k, \varphi_k) = (\partial_\varphi
f)(\theta_k, \varphi_k) = 0$ holds. Then, we obtain the following equations from
the partial derivatives of $f$ with respect to $\theta$ and $\phi$.
\begin{align}
(\partial_\theta f)(\theta_k, \varphi_k)
&= \frac{2\sin \theta_k}{
    \cos^3 \theta_k \cos^{2k} \varphi_k (1 - \cos \varphi_k)^2}
- \frac{2\cos \theta_k}{\sin^3 \theta_k (1 - \cos \varphi_k)^2} \nonumber \\
&= \frac{
    2\tan^2 \theta_k \sin^2 \theta_k - 2\cos^2 \theta_k \cos^{2k} \varphi_k
}{
    \sin^2 \theta_k \cos^2 \theta_k \tan \theta_k \cos^{2k}\varphi_k (1-\cos
    \varphi_k)^2
} \nonumber \\
&= \frac{
    2(\tan^2 \theta_k - \cos^{k} \varphi_k)(\tan^2 \theta_k + \cos^k \varphi_k)
}{
    \sin^2 \theta_k \tan \theta_k \cos^{2k}\varphi_k (1-\cos \varphi_k)^2
} = 0 \label{eq:partial_theta} \\
(\partial_\varphi f)(\theta_k, \varphi_k)
&= \frac{2k \sin \varphi_k}{
    \cos^2 \theta_k \cos^{2k+1} \varphi_k (1 - \cos \varphi_k)^2} \nonumber \\
&\phantom{=\,} - \frac{2 \sin \varphi_k}{(1 - \cos \varphi_k)^3} 
\left( 
    \frac{1}{\sin^2 \theta_k} + \frac{1}{\cos^2 \theta_k \cos^{2k} \varphi_k}
\right) = 0 \label{eq:partial_phi}
\end{align}
Then, from Eq.s~\eqref{eq:partial_theta} and \eqref{eq:partial_phi}, we derive
\begin{equation}
\tan^2 \theta_k = \cos^{k} \varphi_k \label{eq:tan}
\end{equation}
and
\begin{align*}
k &= \frac{\cos^{2k+1} \varphi_k \cos^2 \theta_k}{(1 - \cos \varphi_k)}
\times \left(
    \frac{1}{\sin^2 \theta_k} + \frac{1}{\cos^{2k} \varphi_k \cos^2 \theta_k}
\right),
\end{align*}
respectively.
Combining these equations, we obtain
\begin{align*}
    k &= \frac{\cos^{2k+1} \varphi_k \cos^2 \theta_k}{(1 - \cos \varphi_k)}
    \times \left(
        \frac{1}{\sin^2 \theta_k}
            + \frac{1}{\cos^{2k} \varphi_k \cos^2 \theta_k}
    \right) \\
    &= \frac{\cos \varphi_k \tan^4 \theta_k \cos^2 \theta_k}{
        (1 - \cos \varphi_k)}
    \times \left(
        \frac{1}{\sin^2 \theta_k}
        + \frac{1}{\tan^4 \theta_k \cos^2 \theta_k}
    \right) &(\because Eq.~\eqref{eq:tan}) \\
    &= \frac{\cos \varphi_k \sin^4 \theta_k }{
        (1 - \cos \varphi_k)\cos^2 \theta_k}
    \times \left(\frac{\sin^2 \theta_k+\cos^2 \theta_k}{\sin^4 \theta_k}\right)
    &\\
    &= \frac{\cos \varphi_k}{(1 - \cos \varphi_k)} \times \sec^2 \theta_k
    = \frac{\cos \varphi_k}{(1 - \cos \varphi_k)}
    \times \left(\tan^2 \theta_k + 1\right) \\
    &= \frac{\cos \varphi_k}{(1 - \cos \varphi_k)}
    \times \left(\cos^k \varphi_k + 1\right)
    &(\because Eq.~\eqref{eq:tan})\\
    &= \cos^{k+1} \varphi_k + (k+1) \cos \varphi_k.
\end{align*}
Therefore, if $\omega$ is a unique positive solution of an
equation~$\omega^{k+1} + (k+1) \omega - k = 0$,
then the optimal parameters are
$\theta_k = \tan^{-1}\sqrt{\omega^k}\text{ and } \varphi_k = \cos^{-1}\omega.$
\end{proof}

\subsection{Closure Properties of QFLs}
\label{sec:qfl_operation}

In our framework, we consider Boolean operation over $\QFL$, and distinguish the
closure and error-bound properties of each operation. The closure operations are
considered for each language class: $\MOQFL$, $\MMQFL$, $\MMQFL_\subend$ and
$\MMQFL_\subcoend$. These also include the basic constructions of each language.
Unlike other classes that are defined based on cut-points as mentioned in
previous literature\cite{BrodskyP02,AmbainisF98}, the properties of these
classes have only been indirectly discussed. We examine properties of one-sided
error-bounded languages, including closure properties for operations, to clarify
how our framework supports the operations.

From an MO- and MM-QFA, we can obtain their complement QFAs by exchanging the
set of accepting states and the set of rejecting states. We establish the
following properties from the construction that transforms an end-decisive
MM-QFA from a co-end-decisive MM-QFA and vice versa.

\begin{property}\label{property:coMOQFL}
$\coMOQFL$ and $\coMMQFL$ are
the class of languages recognized by MO-QFAs and MM-QFAs with a bounded
\emph{positive} one-sided error, respectively.
\end{property}

The proof of Property~\ref{property:coMOQFL} follows from the closure property
of stochastic languages of QFAs under complement.

\begin{proof}
Let $L$ be a $\coMOQFL$. By definition, $\overline{L}$ is a $\MOQFL$, meaning
there exists an MO-QFA~$M$ and a margin~$\epsilon > 0$ such that $w \in
\overline{L} \equiv f_M(w) = 1$ and $w \notin \overline{L} \equiv f_M(w) < 1 -
\epsilon$. Given that stochastic languages are closed under complement, we can
construct another MO-QFA~$N$ such that $f_M(w) = 1 - f_N(w)$. Therefore, $w
\notin L \equiv f_N(w) = 0$ and $w \in L \equiv f_N(w) > \epsilon$. Thus,
MO-QFA~$N$ recognizes $L$ with a bounded positive one-sided error. The proofs
for the inverse direction are omitted for brevity. The proof for $\coMMQFL$ is
straightforward and follows similarly.
\end{proof}

\begin{property}\label{property:coMMQFL_end}
$\coMMQFL_\subend$ and $\coMMQFL_\subcoend$ are the class of languages
recognized by \emph{co-end-decisive} and \emph{end-decisive} MM-QFAs with a
bounded \emph{positive} one-sided error, respectively. \end{property}
\begin{proof} Property~\ref{property:coMMQFL_end} directly results from the way
in which the complement of an MM-QFA is constructed, which involves swapping the
end-decisiveness and co-end-decisiveness. In the proof of
Property~\ref{property:coMOQFL}, if $M$ is an end-decisive MM-QFA, then the
corresponding MM-QFA~$N$ can be co-end-decisive, and vice versa.
\end{proof}

We show that the operations are closed under stochastic language and correlate
with the properties of QFLs.

\begin{theorem}\label{theorem:closedness_intersection}
Each class~$\MOQFL$, $\MMQFL$, $\MMQFL_\subend$ and $\MMQFL_\subcoend$ of
languages is closed under intersection.
\end{theorem}

\begin{proof}
The closure property of each language under intersection follows from the
closure property of their corresponding QFAs under linear combination.
\end{proof}

\begin{theorem}\label{theorem:closedness_union}
$\MOQFL$ and $\MMQFL_\subcoend$ are closed under union.
\end{theorem}

\begin{proof}
Let $L_1$ and $L_2$ be languages in $\MOQFL$. By
Property~\ref{property:coMOQFL}, there exist MO-QFAs~$M_1$ and $M_2$ recognizing
$\overline{L_1}$ and $\overline{L_2}$ with a positive one-sided bounded error.
Hence, there exists an MO-QFA~$M$ satisfying $f_M(w) = (f_{M_1} \otimes
f_{M_2})(w)$ due to the closure property of MO-QFAs under Hadamard product. $M$
recognizes $\overline{L_1} \cap \overline{L_2}$ with a positive one-sided error,
and by Property~\ref{property:coMOQFL} again, $\overline{\overline{L_1} \cap
\overline{L_2}} = L_1 \cup L_2$ is a $\MOQFL$~language. For the closure property
of $\MMQFL_\subcoend$, we can utilize Property~\ref{property:coMMQFL_end}
instead of Property~\ref{property:coMOQFL}, and the closure property of
end-decisive MM-QFAs with regard to the Hadamard product.
\end{proof}

The next statement directly follows the fact that there exists a co-end-decisive
MM-QFA for each unary finite language. The corollary leverages the closure
property of $\MMQFL_\subcoend$ with regard to its union and the concept of
singleton language construction from Theorem~\ref{theorem:equ_construction}.

\begin{corollary}
Every unary finite language is in $\MMQFL_\subcoend$.
\end{corollary}

Table~\ref{tab:qfl_closure_property} summarizes the upper bounds of operational
complexities as well as the complexities of inverse homomorphism and word
quotient---both of which can be easily established from the their
constructions~\cite{MooreC00,BrodskyP02}.

\begin{table}[!ht]
\caption{
Upper bounds of operation complexities for QFL classes. Some of operations in
this table are not yet supported by our framework. A dash~(-) indicates that the
upper bound of the operator is the same as the nearest non-empty row above.
}
\label{tab:qfl_closure_property}
\centering
\begin{tabular}{lcccc}
\toprule
QFL Class
& Union
& Intersection
& Inverse Homo.~($h^{-1}$)
& Word Quotient~($w \backslash \cdot$) \\
\cmidrule(r){1-1}
\cmidrule(l){2-5}
$\MOQFL$
& $nm$ & $n + m$ & $n$ & $n$ \\
\cline{4-5}
$\MMQFL$
& & $n + m$
& \multicolumn{1}{|c|}{}
& \multicolumn{1}{c|}{} \\
$\MMQFL_\subend$
& & $n+m$
& \multicolumn{1}{|c|}{$n + c(h) \cdot n_\halt$}
& \multicolumn{1}{c|}{$n + |w| \cdot n_\halt$}
\\
$\MMQFL_\subcoend$
& $nm$ & $n+m$
& \multicolumn{1}{|c|}{}
& \multicolumn{1}{c|}{} \\
\cline{4-5}
\bottomrule
\multicolumn{5}{l}{\footnotesize
$c(h)$ and $n_\halt$ denote
the $\max |h(\sigma)|$ and $|Q_\halt|$, respectively.
}
\end{tabular}
\addtolength{\tabcolsep}{-3pt}
\end{table}
    
\subsection{Density Mapping from QFA States to Qubit Basis Vectors}

After a user composes QFAs for their QFLs using either simple constructions or
language operations, the next step is to transpile these QFAs into quantum
circuits. In the implementation of a QFA in a quantum computer with $n$~qubits,
each state of the QFA is represented with $n$-bits as one of $2^n$ basis vectors
and the transitions are realized through a combination of quantum gates.
Additionally, in the current era of noisy intermediate-scale quantum
computing~\cite{Preskill18}, we need to apply error mitigation methods that can
compensate for possible errors. 
\begin{figure*}[htb]
    \centering
        \includegraphics[width=.9\textwidth]{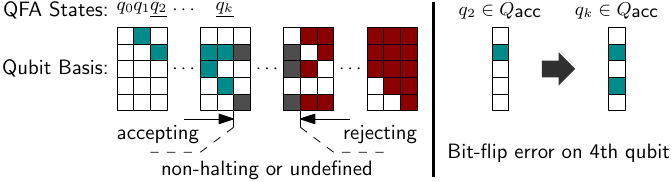}
    \caption{
        The expected error reduction of the proposed $\mathcal{D}$-mapping.
        The accepting state~$q_1$ is mapped to the qubit basis~$\ket{10000}$.
        If a bit-flip error occurs on the fourth qubit,
        then the measured qubit basis is~$\ket{10010}$
        that describes the state~$q_n$.
        It does not effect whether or not the string is accepted
        because both $q_1$ and $q_k$ are accepting.
    }
    \label{fig:mapping}
\end{figure*}

We propose a technique called \emph{density ($\mathcal{D}$-) mapping} which is
applied before the transpilation stage of the quantum circuit. Rather than
mapping QFA states to qubit basis vectors in an arbitrary manner, our
$\mathcal{D}$-mapping reorders these vectors based on the number of qubits with
value~$1$. Then we assign the accepting and rejecting states to the basis
vectors with a low and high number of qubits having the value~$1$, respectively.

$\mathcal{D}$-mapping maximizes the localization of accepting and rejecting
states, thereby reducing the probability of bit-flip errors. These errors can
arise from a decoherence or depolarization, and causes an accepting state to
change into a rejecting state, and vice versa. Additionally, the basis vectors
representing the accepting and rejecting states are separated by basis mapped to
non-final states and basis that are not mapped to any states. This method helps
to isolate the errors and reduce possible error situations.
Fig.~\ref{fig:mapping} conceptually illustrates how our proposal reduces a
bit-flip error from altering the acceptance of a QFA.

\section{Experimental Results and Analysis}
\label{sec:experiments}

We demonstrate the effectiveness of our $\mathcal{D}$-mapping with experiments
using our framework. Our experiments measure how accurately quantum circuits
simulate an operation of $\MOD_p$ of $O(\log{p})$ states and $\EQU_k$ of $4$
states by default construction. We compare the difference between ideal and
experimental acceptance probabilities with and without the
$\mathcal{D}$-mapping. We take 100,000 shots and measure the number of
acceptances and rejections for each string.

We use IBM's Falcon r5.11 quantum processor for real quantum experiments. When
simulating the quantum circuits with controlled error rates, we utilize Qiskit
simulators\footnote{\url{https://www.ibm.com/quantum/qiskit}}. The Qiskit
simulator is configured to use the same basis gates and topological structure as
the Falcon processor. We also set the error rates to follow the Falcon
processor, except for the two types of errors---gate and readout errors. A
mirroring factor~$\eta_g$ is applied to each error rate of gate~$G$ in the
processor. The error rate of $G$ in the simulator is calculated as follows:
\[
    \text{(Error rate of $G$ in the simulator)} 
    = \eta_g \times \text{(Average error rate of $G$ in the processor)}.
\]
Readout errors are adjusted using the mirroring factor~$\eta_r$ and the readout
error in the simulator is calculated similarly.

\subsection{Difficulty of QFA simulation}
\label{sec:real_device}

\begin{figure*}[h!tb]
\centering
\includegraphics[width=0.9\textwidth]{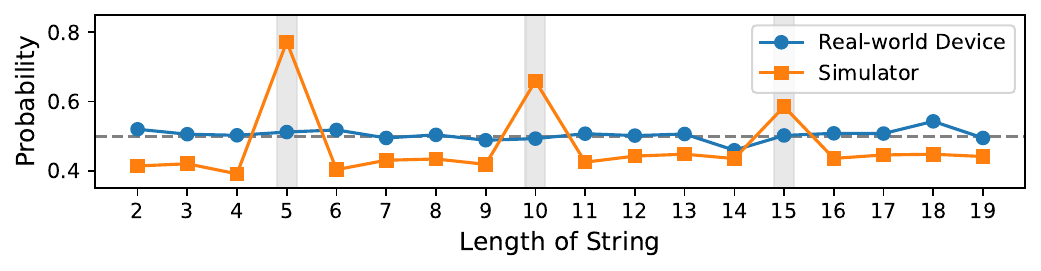}
\caption{
The experimental results for the language~$\MOD_5$. Blue circles and orange
squares indicate the accepting probability of strings, which are simulated on a
real-world quantum computer and a simulator, respectively. The dashed line
represents the 0.5 mark for random guessing. We set the mirroring
factors~$\eta_g$ and~$\eta_r$ as 0.5\%. In the case of an ideal computer with no
errors, the accepting probabilities highlighted by gray boxes are~1.
}
\label{fig:falcon}
\end{figure*}

We measured the $\MOD_5$ language using the Falcon processor, as depicted in
Fig.~\ref{fig:falcon}. This measurement was to demonstrate the current
capabilities and state of quantum computers. Note that the automaton of $\MOD_5$
accepts strings of length divisible by $5$ with probability~$1$. Otherwise, it
accepts the string with a probability less than 0.875. However, we can see that
the measurements for the real-world computer accept strings with a probability
close to 0.5. We analyze the low accuracy attributed to the large size of the
resulting quantum circuit, exceeding 1,000 gates for even a length 2 input.

We also notice despite the small error rates of the simulator, the output of the
simulator becomes inaccurate as the length of the string becomes longer. The
inaccuracy is caused by the size of the circuit that grows linearly to the
length of the input string. The accumulated errors eventually dominate and
reduce the output closer to random guessing.

Our experiments from this point on are conducted using a quantum simulator to
focus on the effectiveness of $\mathcal{D}$-mapping. We use error mirroring
rates designed to reflect the capabilities of near-future quantum computers with
error suppression and mitigation.

\subsection{Effectiveness of the Density Mapping on Simulators}
\label{aer_experiment}

\begin{figure*}[h!tb]
\centering
\includegraphics[width=0.9\textwidth]{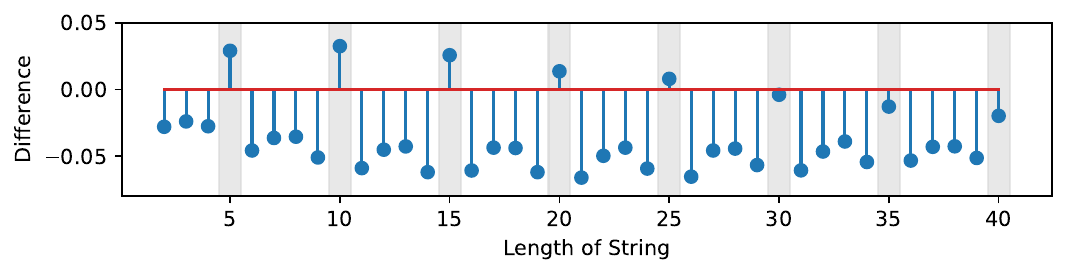}
\caption{
The difference in accepting probability from using the naive mapping to using
$\mathcal{D}$-mapping on $\MOD_5$. We set the mirroring factors~$\eta_g$
and~$\eta_r$ as 0.5\%. The blue dots above the red line represent the increase
in accepting probability of $\mathcal{D}$-mapping compared to the naive mapping.
Conversely, the blue dots below the red line represent the decrease in accepting
probability. The strings highlighted with gray boxes should exhibit an increased
accepting probability towards the ideal case, while other strings should show a
decrease.
}
\label{fig:comparison_mapping}
\end{figure*}

We demonstrate the usefulness of our $\mathcal{D}$-mapping for emulating QFAs.
We evaluate the accepting probability of input strings with the automaton of
$\MOD_5$ implemented by quantum circuits. Fig.~\ref{fig:comparison_mapping}
compares the experimental results for two quantum circuits recognizing $\MOD_5$,
one utilizes our mapping and the other does not.

The experimental results confirm that $\mathcal{D}$-mapping increases the
accuracy of the simulation for most cases. The mapping increases the accepting
probability of strings in $\MOD_5$ up to length~$30$ and decreases the accepting
probability of strings that are not in $\MOD_5$. The effectiveness of
$\mathcal{D}$-mapping diminishes with longer strings. Both mapping methods are
increasingly affected by accumulating errors as discussed in
Section~\ref{sec:real_device}, resulting in similar outcomes.

\subsection{Effectiveness of the Density Mapping in Various QFAs}

Table~\ref{tab:effectiveness_mapping} presents the mean absolute percentage
error (MAPE) of the experimental results for $\MOD_p$ and $\EQU_k$
implementations using the naive mapping and $\mathcal{D}$-mapping. Each cell
denotes the dissimilarity between ideal probabilities~$P(w_i)$ and
measured~$\overline{P}(w_i)$ for inputs~$w_1, \ldots, w_n$, and a simulation
with a lower MAPE indicates higher accuracy. MAPE is calculated as follows:
\[
    \MAPE = \frac{1}{n} \sum_{i = 1}^{k}
    \left|
    \frac{P(w_i) - \overline{P}(w_i)}{P(w_i)}
    \right|
    \times 100 (\%).
\]
We use strings with length from $2$ to $40$ to compute the MAPE for each
language. Note that the maximum achievable value of MAPE varies for each
language, hence we cannot directly compare simulation accuracy between different
languages using MAPE. For instance, consider the following cases: (1)~$\eta_g =
0.1 \%$ and $\eta_r = 0.1 \%$; and (2)~$\eta_g = 1 \%$ and $\eta_r = 1 \%$,
where the latter case has greater error rate. Nevertheless, we use the according
formulation to shows the difference between probabilities and its difference.

\begin{table*}[h!]
\caption{
The dissimilarity between the ideal results and the results of a simulator with
naive and density mapping methods with mirroring factors (M.F.)~$\eta_g$ and
$\eta_r$. Each value in the table represents the value of MAPE (\%), which
indicates the average dissimilarity in acceptance rates of strings.
}
\label{tab:effectiveness_mapping}
\centering
\begin{tabular}{ccrccccccrcccccc}\toprule
\multicolumn{2}{c}{M.F.~(\%)} &~~~&
\multicolumn{6}{c}{Naive Mapping} &~~~&
\multicolumn{6}{c}{Density Mapping} \\
\cmidrule{1-2}
\cmidrule{4-9}
\cmidrule{11-16}
\multirow{2}{*}{$\eta_g$} & \multirow{2}{*}{$\eta_r$} && 
\multicolumn{3}{c}{$\MOD_p$}&
\multicolumn{3}{c}{$\EQU_n$}&& 
\multicolumn{3}{c}{$\MOD_p$}&
\multicolumn{3}{c}{$\EQU_n$}\\
&&&$p=3$&5&7 &$n=3$&5&7
&&3&5&7 &3&5&7\\
\midrule
 &0.1  && 2.49 & 1.64 & 1.35 & 0.07 & 0.08 & 0.07 
       && 1.43 & 1.32 & 1.35 & 0.06 & 0.07 & 0.07 \\
0.1&0.5&& 2.90 & 1.35 & 1.25 & 0.09 & 0.09 & 0.10 
       && 1.44 & 1.17 & 1.65 & 0.09 & 0.10 & 0.09 \\
 & 1   && 2.82 & 1.85 & 1.43 & 0.10 & 0.11 & 0.08 
       && 1.92 & 1.21 & 1.66 & 0.13 & 0.11 & 0.10 \\
\midrule
 &0.1  && 12.76 & 7.44 & 5.80 & 0.10 & 0.09 & 0.11 
       &&  6.13 & 4.85 & 6.22 & 0.11 & 0.11 & 0.11 \\
0.5&0.5&& 13.08 & 7.16 & 6.01 & 0.14 & 0.17 & 0.13 
       &&  6.24 & 4.64 & 6.04 & 0.14 & 0.13 & 0.16\\
 & 1   && 12.86 & 7.10 & 5.87 & 0.19 & 0.17 & 0.20 
       &&  6.08 & 4.85 & 6.25 & 0.18 & 0.15 & 0.18 \\
\midrule
 &0.1  && 24.18 & 13.83 & 11.02 & 0.19 & 0.21 &  0.18 
       && 11.66 &  8.39 & 10.94 & 0.19 & 0.19 &  0.19 \\
1&0.5  && 24.26 & 13.73 & 10.91 & 0.23 & 0.20 & 0.23 
       && 11.37 &  8.28 & 11.21 & 0.22 & 0.18 & 0.22 \\
 & 1   && 24.46 & 13.94 & 11.18 & 0.25 & 0.23 & 0.29 
       && 11.31 &  8.36 & 11.15 & 0.24 & 0.26 & 0.27 \\
\bottomrule
\end{tabular}
\end{table*}

The experimental results indicate that $\mathcal{D}$-mapping is effective for
QFAs of languages~$\MOD_3$, $\MOD_5$, but not for $\MOD_7$.
$\mathcal{D}$-mapping reduces the difference between the accepting probabilities
of ideal results and simulation for $\MOD_3$ to $57.43\% (1.43 / 2.49)$ of the
former case and $46.24\% (11.31/24.46)$ of the latter case. Similarly, for
$\MOD_5$, $\mathcal{D}$-mapping reduces the difference to $80.49\%$ and
$59.97\%$ of the former and latter case, respectively.

Note that the number of quantum circuit for $\MOD_p$ increases with the growth
of $p$. We observe that the effect of $\mathcal{D}$-mapping is not evident for
$\MOD_7$, as the simulation remains inaccurate regardless of the mapping used.
This emphasizes the challenge posed by the large size of $\MOD_7$ QFAs and the
resulting quantum circuits, which involve a significant number of gates.

We also examine the impact of $\mathcal{D}$-mapping on QFAs of $\EQU_n$,
although its effect is less pronounced compared to $\MOD_p$. Note that QFAs of
$\EQU_n$ are MM-QFAs. Unlike MO-QFAs for $\MOD_p$, simulations of MM-QFAs have
the possibility of terminating before reading all symbols in each string. We
suggest that the simulation for MM-QFAs receives less effect from gate errors
because they may ignore trailing gates, thereby resulting in the reduced
effectiveness of $\mathcal{D}$-mapping.

The overall experimental results reveal that gate error rates dominate
simulation accuracy for QFAs, and our experiments do not show a clear relation
between readout error rates and simulation accuracy. This is primarily due to
the fact that the quantum circuits for QFAs have few measurement gates.

\section{Conclusions and Future Work}
\label{sec:conclusion}

MO-QFA and MM-QFA are the two most simplest quantum computational models.
Despite their simple structure, designing these models is complex because their
transitions must satisfy the unitary constraints. Verification is also
challenging due to the low accuracy of current quantum devices. We have
developed the framework to facilitate easy construction and accurate simulation
of QFAs. For the development of the framework, we have 
\begin{itemize}
\item composed a novel construction for $\EQU_k$,
\item demonstrated the closure properties of $\MOQFL$, $\MMQFL$, $\MMQFL_\subend$
and $\MMQFL_\subcoend$,
\item introduced $\mathcal{D}$-mapping to improve the accuracy of experiments on
real-world devices,
\item showcased the basic functionalities of our framework.
\end{itemize}

Our experiments on real devices highlight a significant limitation of current
quantum processors---the QFA simulations on these processors are close to the
random guessing due to their low accuracy. On the other hand, the experimental
results confirm that our density mapping increases the accuracy of the QFA
simulations.

A major limitation of our framework is its inability to construct QFAs from a
human-readable format while most FA frameworks support a direct FA construction
from regular expressions. Furthermore, manual construction remains inevitable
for some languages because some types of QFLs cannot be obtained by composing
$\MOD_p$ and $\EQU_k$. We aim to characterize the class of languages by
identifying which languages and operations can generate these classes. We will
then integrate these languages and operations into our framework and enable
users to construct a broader variety of languages more easily.

Our density mapping technique reduces simulation errors by correlating each QFA
state with an appropriate dimension of a quantum circuit. The accuracy in
quantum circuits can also be increased by reducing the number of
gates~\cite{BirkanSONY21} or by assigning more physical qubits to represent a
single logical qubit~\cite{GoshFG12}. The interaction between the density
mapping and these existing methods may affect their overall effectiveness. Thus,
it is another future work to integrate our density mapping with other techniques
for improving the overall performance. 

\bibliography{main}

\end{document}